\documentclass[letterpaper]{article} 
\usepackage{color}
\usepackage{amsmath}
\usepackage{natbib}  
\usepackage{hyperref}

\usepackage{booktabs} 
\usepackage{makecell}
\usepackage{array}
\usepackage{adjustbox}
\usepackage{aaai2026}  
\usepackage{times}  
\usepackage{helvet}  
\usepackage{xspace}
\usepackage{bibentry}
\usepackage{multirow}  
\usepackage{graphicx}  
\usepackage{courier}  
\usepackage{pifont}
\usepackage{amssymb}
\usepackage{caption} 
\usepackage{algorithm}
\usepackage{algorithmic}

\usepackage{newfloat}
\usepackage{listings}

\DeclareCaptionStyle{ruled}{labelfont=normalfont,labelsep=colon,strut=off} 
\floatstyle{ruled}
\newfloat{listing}{tb}{lst}{}
\floatname{listing}{Listing}
\newcommand{\thename}{\textit{ReconDreamer-RL}\xspace}

\title{ReconDreamer-RL: Enhancing Reinforcement Learning via Diffusion-based Scene Reconstruction}
\author{
    Chaojun Ni\textsuperscript{\rm \ 1, 2}\equalcontrib 
    \hspace{1em}
    Guosheng Zhao\textsuperscript{\rm \ 1, 3}\equalcontrib
    \hspace{1em}
    Xiaofeng Wang\textsuperscript{\rm \ 1}\equalcontrib
    \hspace{1em}
    Zheng Zhu\textsuperscript{\rm \ 1}\equalcontrib\coo\\
    \hspace{1em}
    Wenkang Qin\textsuperscript{\rm \ 1}
    \hspace{1em}
    Xinze Chen\textsuperscript{\rm \ 1}
    \hspace{1em}
    Guanghong Jia\textsuperscript{\rm  \ 4}
    \hspace{1em}
    Guan Huang\textsuperscript{\rm \ 1}
    \hspace{1em}
    Wenjun Mei\textsuperscript{\rm \ 2}\footnotemark[2]
}
\affiliations{
    \textsuperscript{\rm 1}GigaAI
    \hspace{1em}
    \textsuperscript{\rm 2}Peking University
    \hspace{1em}
    \textsuperscript{\rm 3}CASIA
    \hspace{1em}
    \textsuperscript{\rm 4}Tsinghua University\\
    \small{Project Page: \url{https://ReconDreamer-RL.github.io}}
}

\hypersetup{
    linkbordercolor=white,   
    pdfborder={1 1 1}        
}

\begin{document}

\maketitle

\begin{abstract}
Reinforcement learning for training end-to-end autonomous driving models in closed-loop simulations is gaining growing attention. However, most simulation environments differ significantly from real-world conditions, creating a substantial simulation-to-reality (sim2real) gap. To bridge this gap, some approaches utilize scene reconstruction techniques to create photorealistic environments as a simulator. While this improves realistic sensor simulation, these methods are inherently constrained by the distribution of the training data, making it difficult to render high-quality sensor data for novel trajectories or corner case scenarios. Therefore, we propose \textit{ReconDreamer-RL}, a framework designed to integrate video diffusion priors into scene reconstruction to aid reinforcement learning, thereby enhancing end-to-end autonomous driving training. Specifically, in \textit{ReconDreamer-RL}, we introduce ReconSimulator, which combines the video diffusion prior for appearance modeling and incorporates a kinematic model for physical modeling, thereby reconstructing driving scenarios from real-world data. This narrows the sim2real gap for closed-loop evaluation and reinforcement learning. To cover more corner-case scenarios, we introduce the Dynamic Adversary Agent (DAA), which adjusts the trajectories of surrounding vehicles relative to the ego vehicle, autonomously generating corner-case traffic scenarios (e.g., cut-in). Finally, the Cousin Trajectory Generator (CTG) is proposed to address the issue of training data distribution, which is often biased toward simple straight-line movements. Experiments show that \textit{ReconDreamer-RL} improves end-to-end autonomous driving training, outperforming imitation learning methods with a 5$\times$ reduction in the Collision Ratio.

\end{abstract}

%

\section{Introduction}

End-to-end training of autonomous driving models~\cite{diffusionad,diffusiondrive,Hydra,sparsedrive,Para,GenDrive,jia2024bench2drive} through reinforcement learning in closed-loop simulation environments attracts growing interest. Compared to imitation learning, which relies solely on expert-collected demonstrations, closed-loop reinforcement learning enables the model to interact with the environment, enhancing its robustness and adaptability across diverse scenarios.

Despite these advantages, existing methods still face significant challenges. One key issue is creating realistic driving environments capable of effective interactions with autonomous driving policies. Game-engine-based simulators~\cite{garchingsim,CarMaker} lack sensor-level realism, whereas real-world closed-loop training is costly and risky. To overcome these limitations, recent approaches~\cite{rad,mars,unisim} employ scene reconstruction methods to build photorealistic digital twins of real-world scenarios. However, these reconstruction-based methods are constrained by their training data, typically generating high-quality sensor outputs only within regions covered by recorded camera trajectories. Additionally, these methods fail to adequately account for corner cases such as sudden braking, as such rare behaviors are typically absent from the reconstruction data.

In this paper, we introduce \textit{ReconDreamer-RL}, a framework that integrates the video diffusion priors~\cite{drivedreamer,drivedreamer2} into scene reconstruction to enhance end-to-end autonomous driving training by reinforcement learning. The framework consists of three components: the ReconSimulator, the Dynamic Adversary Agent (DAA), and the Cousin Trajectory Generator (CTG). \textit{ReconDreamer-RL} enhances the training process in two key stages: (1) The imitation learning stage uses behavior cloning for plan initialization. (2) The reinforcement learning stage optimizes the policy through closed-loop trial-and-error interactions with the environment. Specifically, ReconSimulator employs 3D Gaussian Splatting (3DGS) to reconstruct the driving scene and incorporates the video diffusion prior for appearance modeling. Meanwhile, a kinematic model is used to ensure the validity of vehicles' trajectories during ego-vehicle movement and scene editing. Then, in the first stage, we propose that DAA enrich corner case scenarios by controlling surrounding vehicles to generate challenging situations such as sudden lane changes. Meanwhile, the CTG  enhances the diversity of sensor-collected data by synthesizing new trajectories from expert trajectories, addressing the data bias toward straight-line driving. In the second stage, the autonomous driving policy is trained within ReconSimulator via reinforcement learning, while the DAA continues to dynamically alter surrounding vehicle trajectories to create corner cases.

Through evaluation in the closed-loop 3DGS environment, we demonstrate that the end-to-end driving policy trained with the \textit{ReconDreamer-RL} framework performs better, especially in corner cases, with a 5$\times$ reduction in Collision Ratio compared to imitation learning methods.

The main contributions of this paper are summarized as follows:
\begin{itemize}

    \item We introduce \textit{ReconDreamer-RL}, the framework that enhances end-to-end autonomous driving training by scene reconstruction with diffusion prior. It uses ReconSimulator to create realistic and freely explorable environments for reinforcement learning by enhancing appearance and physical modeling,  reducing the sim2real gap.

    \item We propose DAA to auto-generate corner case instances to improve scenario coverage. Additionally, the CTG is introduced to address data distribution challenges by enriching sensor-collected expert trajectories, ensuring more balanced and comprehensive training data.
    
    \item We validate the effectiveness of \textit{ReconDreamer-RL} in a closed-loop 3DGS environment. \textit{ReconDreamer-RL} performs stronger in challenging closed-loop evaluations, achieving a 5$\times$ reduction in collision rate.
\end{itemize}

\section{Related Work}

\subsection{Driving Scene Reconstruction with Diffusion Prior} 
3DGS has become a prominent technique in 3D scene reconstruction. Various works~\cite{streetgaussian,pvg,s3gaussian,omnire} have applied 3DGS to autonomous driving scenarios. However, these methods exhibit notable performance degradation when rendering trajectories that deviate from the training distribution. To address this issue, some studies leverage video diffusion models~\cite{sd,svd} to enhance the representation of driving scenes. DriveDreamer4D~\cite{drivedreamer4d} is the first to leverage the video diffusion prior to generate novel trajectory videos for training reconstruction models. ReconDreamer~\cite{recondreamer} further improves reconstruction by integrating the video diffusion model with online artifact correction, enabling better rendering of large maneuvers. ReconDreamer++~\cite{recondreamerplus} enhances rendering quality by reducing domain gaps and refining the ground surface. However, these works focus solely on improving scene representation. In contrast, we leverage the video diffusion prior to improve the simulator's appearance modeling and introduce physical modeling, enhancing end-to-end autonomous driving training by reinforcement learning.

\subsection{End-to-End Autonomous Driving} 
Recent advancements in end-to-end autonomous driving algorithms showcase the immense potential of learning-based planning, where raw sensor inputs are directly mapped to control outputs. UniAD~\cite{uniad} integrates various perception tasks~\cite{caril,ilsurvey,leveragingil,keypoint,pluto}  to enhance planning performance. VAD~\cite{vad} focuses on improving trajectory planning by utilizing compact vectorized scene representations. Furthermore, VADv2~\cite{vadv2} extends this by introducing a framework that models probability distributions over planning vocabularies. However, methods based on imitation learning rely heavily on expert demonstrations and suffer from poor generalization. To address this issue, some approaches~\cite{think2drive,carplanner,nuplan,shalev2016safe,gao2024improved,alphadrive} utilize deep reinforcement learning~\cite{master,alphago,alphafold,dqn,ppo,grpo,wang2025embodiedreamer} for training end-to-end autonomous driving models. Nevertheless, these methods often encounter limitations, either due to simulation environments lacking photorealistic fidelity~\cite{CARLA} or reliance on open-loop data. Recent work, RAD~\cite{rad}, introduces a reinforcement learning framework specifically designed for training autonomous driving agents in photorealistic 3DGS environments. However, RAD~\cite{rad} still suffers from inadequate supervision for policy learning due to limitations in rendering novel views and a lack of corner-case scenarios in reconstructed data, leading to limited generalization and a persistent sim-to-real gap.

\begin{figure*}[t]
    \centering
    \includegraphics[width=1\linewidth]{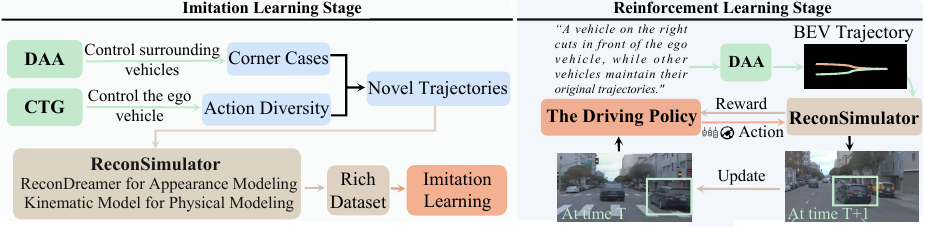}
    \caption{In \textit{ReconDreamer-RL}, ReconSimulator improves appearance modeling by ReconDreamer and incorporates physical modeling to reconstruct driving scenes. In the imitation learning stage, DAA generates corner-case scenario trajectories, while CTG diversifies the ego vehicle's actions and uses ReconSimulator to render sensor data for training the policy. In the reinforcement learning stage, the policy is trained in a closed-loop environment, interacting with DAA-controlled surrounding vehicles.}
     \label{fig:ReconSimulator} 
\end{figure*}

\begin{figure}[h]
    \centering
    \includegraphics[width=1\linewidth]{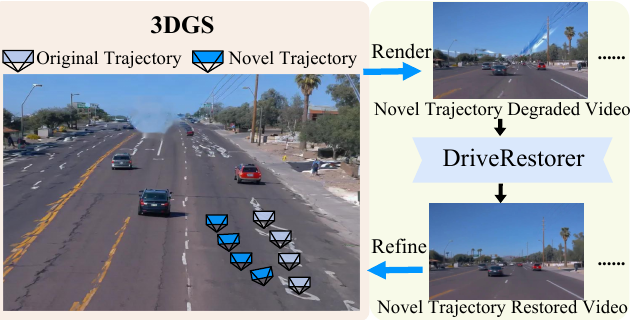}
    \vspace{-1mm}
\caption{The process of integrating the diffusion prior for appearance modeling. During the reconstruction of driving scenes, we first render novel trajectory view videos. These rendered videos are then processed by the DriveRestorer to enhance their visual quality, and the restored results are used to further optimize the reconstruction model. This iterative process continues until the reconstruction model converges.}\
    \label{fig:reconsimulator} 
        \vspace{-4mm}
\end{figure}

\section{Method}
\label{sec:method}
\subsection{Preliminary} 
ReconDreamer~\cite{recondreamer} integrates the video diffusion prior to improve the performance of reconstruction methods in handling large maneuvers. The core of ReconDreamer is DriveRestorer, designed to restore artifacts in rendered videos that arise in novel trajectory views. It is fine-tuned based on the video diffusion models~\cite{drivedreamer,drivedreamer2} and uses a diffusion loss as follows:
\begin{equation}
\mathcal{L}_{\mathcal{R}} = \mathbb{E}_{\boldsymbol{z}, \epsilon \sim \mathcal{N}(0,1), t} \left[ \left\| \epsilon_t - \epsilon_\theta \left( \boldsymbol{z}_t, t, \boldsymbol{c} \right) \right\|_2^2 \right],
\end{equation}
where $\epsilon_t$ is the random noise at time step $t$, $\epsilon_{\theta}$ is the denoising network, $\boldsymbol{z}_t$ is the noisy latent variable, and $\boldsymbol{c}$ represents the control conditions, including the degraded video $\hat{V}_{\text{novel}}$, 3D bounding boxes, and HDMaps.

During the inference stage, DriveRestorer freezes its network parameters and restores novel trajectory renderings. The inference process is expressed as:
\begin{equation}
V_{\text{novel}} = \mathcal{R}(\hat{V}_{\text{novel}},s),
\end{equation}
where $\mathcal{R}$ represents the DriveRestorer model, and $s$ denotes the structural conditions corresponding to the degraded video $\hat{V}_{\text{novel}}$ (such as 3D bounding boxes and HDMaps). In \textit{ReconDreamer-RL}, we leverage DriveRestorer to enhance scene rendering quality, enabling free ego car navigation and scene editing, as detailed in the ReconSimulator Section.

\begin{figure*}[t]
    \centering
    \includegraphics[width=1\linewidth]{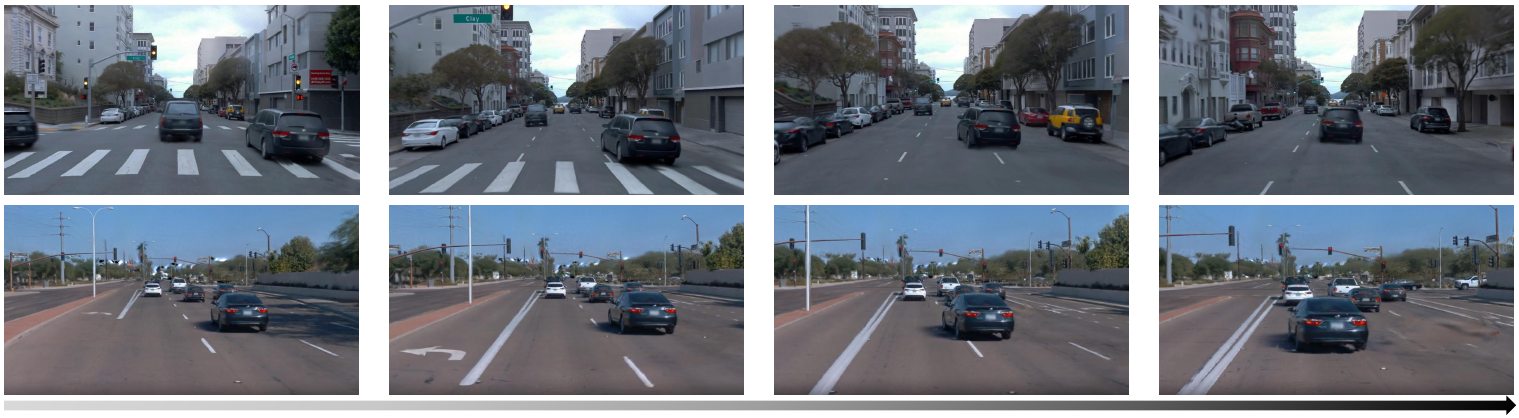}
      \vspace{-2mm}
    \caption{Examples of Dynamic Adversary Agent (DAA) controlling surrounding vehicles to simulate cut-in scenarios.}
    \label{fig:novel_view} 
    \vspace{-4mm}
\end{figure*}

\begin{figure}[h]
    \centering
    \includegraphics[width=1\linewidth]{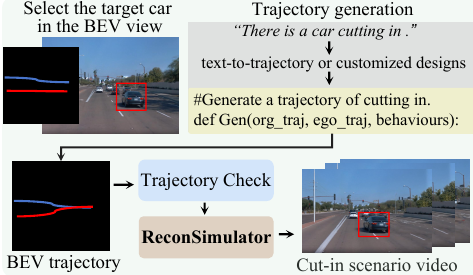}
    \vspace{-1mm}
\caption{The pipeline of the DAA. DAA identifies the target vehicles based on their distances to the ego car from the BEV view, where the blue line represents the ego car’s trajectory and the red line represents the target vehicle. Then, DAA generates novel trajectories based on the specified interactive behavior. The generated trajectories are checked, and feasible ones are rendered using ReconSimulator.}
    \label{fig:DAA} 
        \vspace{-5mm}
\end{figure}

\subsection{Overview of the \textit{ReconDreamer-RL} framework}
Existing end-to-end autonomous driving algorithms are typically trained based on imitation learning, leading to poor generalization. To address this, RAD~\cite{rad} trains the policy in 3DGS-based simulations, which provide photorealistic sensor data. However, it still faces challenges in rendering novel views and covering diverse corner cases.

We introduce the \textit{ReconDreamer-RL} pipeline in Fig.~\ref{fig:ReconSimulator}. ReconSimulator first reconstructs the driving scene, providing high-fidelity sensor data for ego vehicle navigation by integrating video diffusion priors and physical modeling. In the imitation learning stage, DAA generates corner cases for imitation learning, while CTG improves driving behavior through diverse actions. In the reinforcement learning stage, the policy is trained in closed-loop environments, with DAA generating new corner case trajectories to increase difficulty.

\subsection{ReconSimulator}
\label{section:reconsimulator}
The ideal reinforcement learning environment for autonomous driving should minimize the sim2real gap. However, existing 3DGS-based methods merely reconstruct data, failing to ensure realistic appearances along novel trajectories. Meanwhile, it is essential to ensure that the trajectories of all vehicles adhere to physical constraints. Therefore, we propose ReconSimulator to address appearance and physical modeling challenges. In terms of appearance, we integrate the diffusion prior to enhance scene representation for high-quality sensor data rendering on novel trajectories. For physical modeling, kinematic modeling is used to ensure the physical feasibility of vehicle trajectories. Next, we delve into the details of the appearance and physical modeling.

\noindent\textbf{Appearance Modeling.} To ensure high-quality rendering and scene editing, we first use 3DGS to reconstruct the scene and render novel trajectories. Then, DriveRestorer~\cite{recondreamer} mitigates artifacts in the rendered videos, and the results are used to fine-tune the reconstruction model. This enables the final model to produce high-quality rendering from diverse viewpoints. The overall process is illustrated in Fig.~\ref{fig:reconsimulator}. Meanwhile, we model the static background and moving vehicles separately, which enables the modification of trajectories and appearances for all vehicles. The background model is represented by \( G_{\text{Background}, w} \) in the world coordinate system. For each moving object \( v \), represented by Gaussians \( G_{\text{Rigid}, l}^{v} \) in the local coordinate system \( l \), these Gaussians must be transformed into the background coordinate system during the rendering process:
\begin{equation}
G_{\text{Rigid}, w}^{v}(t) = M_t^v \cdot G_{\text{Rigid}, l}^{v} + S_t^v,
\end{equation}
where \( M_t^v \) and \( S_t^v \) represent the rotation matrix and translation vector that describe the pose of object \( v \) at time \( t \).
\noindent\textbf{Physical Modeling.} To maintain the authenticity of vehicle trajectories, a kinematic model is used to govern the vehicle's motion. Specifically, the vehicle's pose at time \( t \) in the world coordinate system is \( W_t = [R_t \mid P_t] \in SE(3) \), where \( R_t \) is the rotation matrix and \( P_t \) is the position. At each time step, the pose is updated using linear velocity \( v_t \) and steering angle \( \delta_t \) in a kinematic bicycle model:
\begin{equation}
P_{t+1} = P_{t} + v_t \cdot \Delta t \cdot \hat{d}_t,
\end{equation}
where $\hat{d}_t$ is the forward direction vector derived from the rotation matrix $R_t$. The vehicle orientation is updated by applying a rotation about the vertical axis:
\begin{equation}
R_{t+1} = \text{Rot}_z(\Delta \theta_t) \cdot  R_{t},
\end{equation}
where the incremental rotation angle $\Delta \theta_t$ is calculated based on the bicycle model as:
\begin{equation}
\Delta \theta_t = \frac{v_t}{L}\tan(\delta_t)\Delta t.
\end{equation}
The $L$ denotes the wheelbase length of the vehicle. The rotation matrix $\text{Rot}_z(\Delta \theta_t)$ is explicitly expressed as:
\begin{equation}
\text{Rot}_z(\Delta \theta_t) = 
\begin{bmatrix}
\cos(\Delta \theta_t) & \sin(\Delta \theta_t) & 0 \\
-\sin(\Delta \theta_t) & \cos(\Delta \theta_t) & 0 \\
0 & 0 & 1
\end{bmatrix}.
\end{equation}
We define different kinematic parameters for each vehicle category and perform checks when updating the trajectories to ensure the trajectory updates remain physically plausible and within the vehicle's operational constraints, including the maximum steering angle and velocity.

\begin{figure*}[t]
    \centering
    \includegraphics[width=1\linewidth]{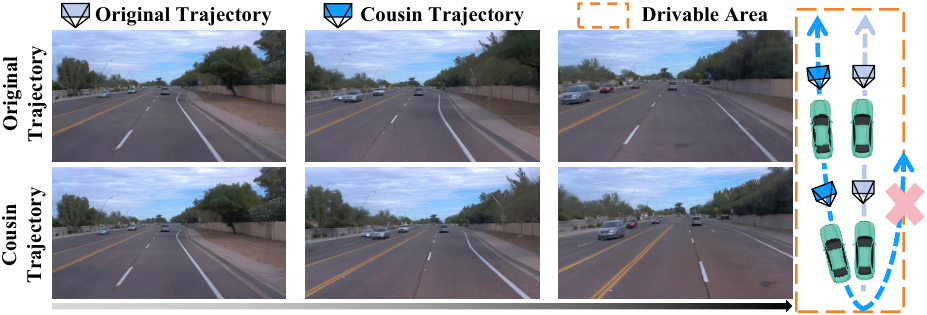}
    \caption{Cousin Trajectory Generator (CTG) generates cousin trajectories and performs trajectory checks to eliminate unreasonable trajectories (e.g., the pink cross marks), and finally renders the corresponding sensor data in the ReconSimulator.}
    \label{fig:CTG} 
\end{figure*}

\begin{figure}[t]
    \centering
    \includegraphics[width=1\linewidth]{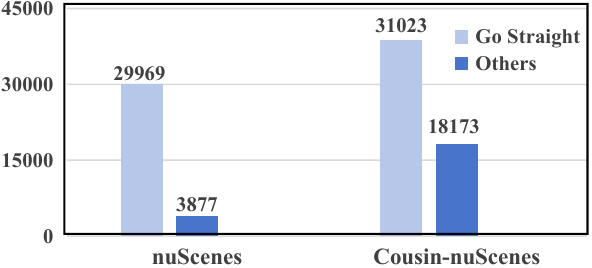}
    \caption{The Cousin-nuScenes dataset created by CTG has an increased variety of actions compared to the nuScenes.}
    \vspace{-4mm}
    \label{fig:rl_2} 
\end{figure}

\subsection{Dynamic Adversary Agent}
Existing  end-to-end autonomous driving models are trained with limited exposure to corner cases. To tackle this, we propose the Dynamic Adversary Agent (DAA), which generates realistic and diverse corner case interactions. The overview of DAA is provided in Fig.~\ref{fig:DAA}, and as shown in Fig.~\ref{fig:novel_view}, it can generate complex cut-in scenarios.

DAA first identifies suitable interactive target vehicles from the Bird's-Eye View (BEV) perspective, based on the distances to the ego vehicle and the specified interactive behavior \( \mathcal{B} \). Different interactive behaviors correspond to different distance settings, which refer to traditional traffic flow methods~\cite{kesting2013traffic,treiber2013traffic}. Then, a new trajectory is generated for the selected target vehicle by modifying its original trajectory \( T_\text{target} \) based on both the ego vehicle's trajectory \( T_\text{ego} \) and  \( \mathcal{B} \). The new trajectory \( T^{'}_\text{target} \) can be represented as:
\begin{equation}
\label{equ:1}
T^{'}_\text{target} = f(T_\text{ego}, T_\text{target}, \mathcal{B}),
\end{equation}
where \( f(\cdot) \) represents the trajectory generation function, which can be implemented using methods such as text-to-trajectory~\cite{drivedreamer2} and customized based on specific requirements. Then, the generated trajectory is checked for feasibility. First, the vehicle trajectory \( T'_\text{target} \) should remain within the drivable region, and avoid collisions with other vehicles (collisions with the ego-vehicle are allowed):
\begin{equation}
    \begin{aligned}
        & \|T'_\text{target} - o_j\| \geq d_{\text{min}}, \quad \forall j \in \{1, \dots, M\},
    \end{aligned}
\end{equation}
where \( d_{\text{min}} \) is the minimum distance between different agents \( \{ o_j \}_{j=1}^M \). Then, the trajectory is further checked to meet the constraints of the kinematic model and  converted into the BEV perspective to verify whether it satisfies the interactive behavior \( \mathcal{B} \).

DAA can be used in both imitation learning and reinforcement learning stages. In the first stage, DAA generates corner case scenarios by controlling surrounding vehicles and generating the ego vehicle's trajectory to avoid collisions, using a similar method mentioned above. The ReconSimulator then renders offline autonomous driving data based on the trajectories for imitation learning. In the second stage, the reinforcement learning policy controls the actions of the ego vehicle, while DAA controls the trajectory of the target vehicle in ReconSimulator to create corner cases. To enhance robustness, DAA has a certain probability of fine-tuning the trajectories (e.g., adjusting the target vehicle's speed) instead of directly reusing them from the first stage.

\begin{figure*}[t] 
\centering
\includegraphics[width=1\textwidth]{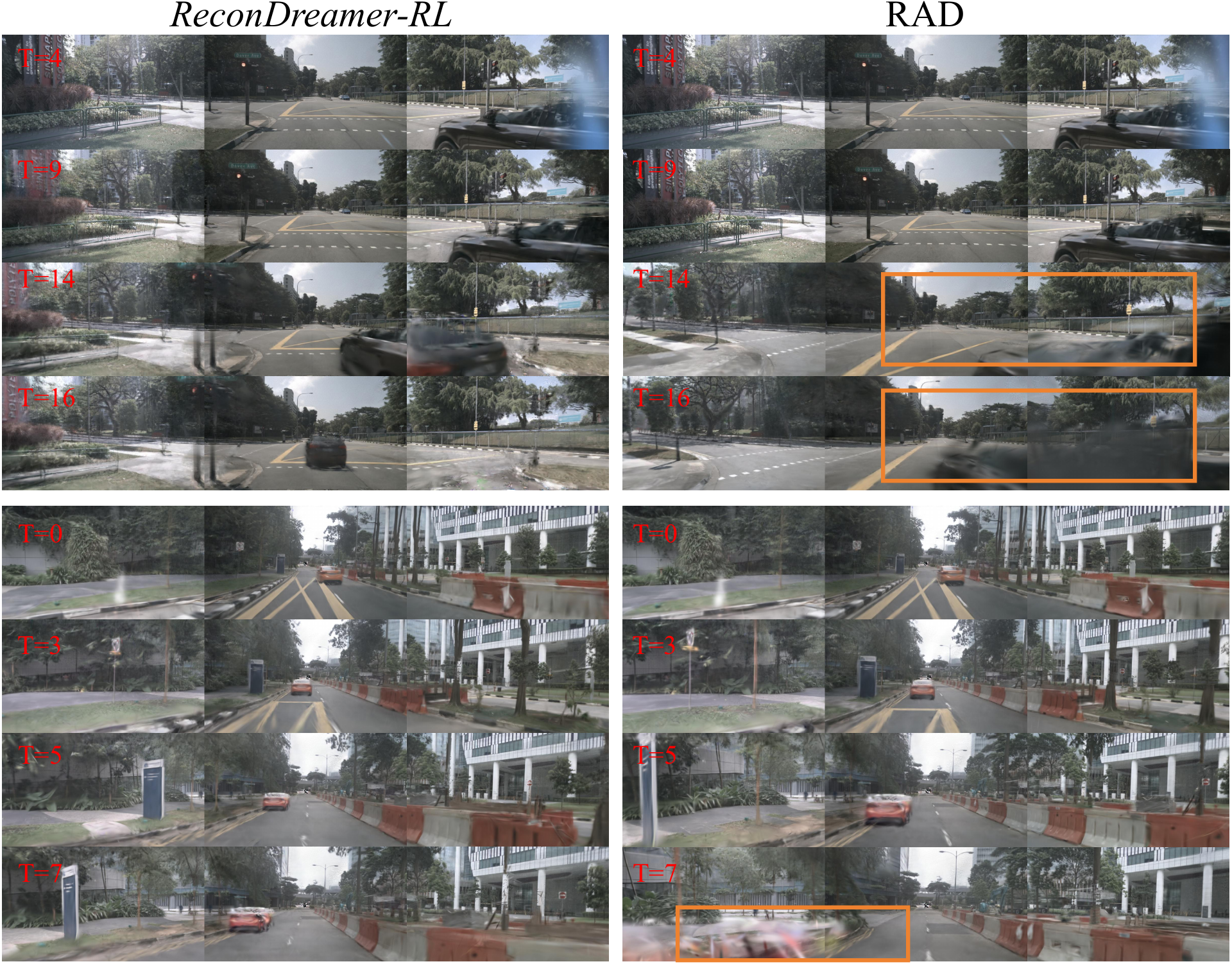}
\vspace{-3mm}
\caption{Comparison of different methods in challenging corner cases, with collisions highlighted by orange boxes.}
\label{fig:visual}
\end{figure*}

\subsection{Cousin Trajectory Generator}
To address cold-start issues in reinforcement learning~\cite{start1,start2}, behavior cloning for pretraining is effective but requires large-scale real-world data, which is often biased toward simple scenarios~\cite{clone_weak1,clone_weak3}. We propose the Cousin Trajectory Generator~(CTG) to enhance action diversity via trajectory extension and interpolation. It is used to create the Cousin-nuScenes dataset with more diverse actions.

For trajectory extension, we generate new trajectories for the ego vehicle, such as lane changes and sharp turns, following a process similar to that described in Equation.~\ref{equ:1}. Subsequently, a trajectory check is conducted to evaluate the validity of the generated trajectories, ensuring they remain physically plausible and adhere to the vehicle's operational limitations, including steering angle and velocity, and verifying collision avoidance from the BEV perspective. Meanwhile, to better leverage expert trajectories from rare scenarios like U-turns, we interpolate the expert data for more detailed driving information. Given expert trajectories \( \mathbf{X}_{\text{ego}} = \{ X_{\text{ego}}^{t_1}, X_{\text{ego}}^{t_2}, \dots, X_{\text{ego}}^{t_n} \} \), where each \( X_{\text{ego}}^{t_i} \) is the ego vehicle's position at time \( t_i \), we perform linear interpolation between consecutive time steps \( t_i \) and \( t_{i+1} \). For an interpolated point \( X_{\text{ego}}^t \) at time \( t \), where \( t_i \leq t \leq t_{i+1} \), the formula is:
\begin{equation}
X_{\text{ego}}^t = X_{\text{ego}}^{t_i} + \frac{t - t_i}{t_{i+1} - t_i} \left( X_{\text{ego}}^{t_{i+1}} - X_{\text{ego}}^{t_i} \right).
\end{equation}
The time \( t \) is expressed as:
\begin{equation}
t = t_i + k \cdot \Delta t, \quad k \in \{1, 2, \dots, m\},
\end{equation}
where \( \Delta t = \frac{t_{i+1} - t_i}{m+1} \) and \( m \) is the number of interpolation points between two time steps. For each interpolated trajectory point \( X_{\text{ego}}^t \), we accordingly adjust the positions of surrounding vehicles to ensure the interpolated trajectories maintain a realistic spatial relationship and interactions. 

After generating the interpolation and extension trajectories, we use the ReconSimulator to obtain the corresponding sensor data, which is then used for imitation training. We provide examples of trajectory extension in Fig.~\ref{fig:CTG}, illustrating two cases of ego-vehicle lane-changing maneuvers. Moreover, as shown in Fig.~\ref{fig:rl_2}, Cousin-nuScenes created by CTG  contains 4$\times$ more non-straight-line driving maneuvers than the nuScenes~\cite{nuscenes} dataset.

\section{Experiments}

\begin{table*}[t]
\centering
\resizebox{1\textwidth}{!}{
\setlength{\tabcolsep}{1em} 
\renewcommand{\arraystretch}{1.0} 
\begin{tabular}{p{13em} c c c c c c c c}
\toprule
Method & CR $\downarrow$ & DCR$\downarrow$ & SCR$\downarrow$ & DR$\downarrow$ & PDR$\downarrow$ & HDR$\downarrow$ \\
\midrule
VAD~\cite{vad}   & 0.386 & 0.234 & 0.152 &  0.163 & 0.103 & 0.060  \\
GenAD~\cite{genad} & 0.333 & 0.190 & 0.143 & 0.146 & 0.093 & 0.053 \\
VADv2~\cite{vadv2} & 0.290 & 0.162 & 0.128 & 0.154 & 0.107 & 0.047 \\
RAD~\cite{rad} & 0.238 & 0.143 & 0.095 &0.084 & 0.057 & 0.027  \\
\textit{ReconDreamer-RL}  & \textbf{0.077} & \textbf{0.048} & \textbf{0.029} &   \textbf{0.040} & \textbf{0.027} & \textbf{0.013} \\
\bottomrule
\end{tabular}%
}
\caption{Comparison of different methods on various metrics.}
\label{tab:comparison}
\end{table*}

\begin{table*}[h]
\centering
\resizebox{1\textwidth}{!}{ 
\setlength{\tabcolsep}{1em} 
\renewcommand{\arraystretch}{1.0} 
\begin{tabular}{>{\centering\arraybackslash}p{5.5em} >{\centering\arraybackslash}p{1.59em} >{\centering\arraybackslash}p{1.59em} c c c c c c}

\toprule
ReconSimulator & DAA & CTG & CR$\downarrow$ & DCR$\downarrow$ & SCR$\downarrow$ & DR$\downarrow$ & PDR$\downarrow$ & HDR$\downarrow$ \\
\midrule
 &&   & 0.238 & 0.143 & 0.095 & 0.084 & 0.057 & 0.027  \\
  $\checkmark$ &&   & 0.172 & 0.103 & 0.069 & 0.073 & 0.053 & 0.020  \\
 $\checkmark$ &$\checkmark$ & & 0.117 & 0.069 & 0.048 & 0.067 & 0.050 & 0.017  \\
$\checkmark$  && $\checkmark$ & 0.143 & 0.086 & 0.057 & 0.053 & 0.040 & 0.013  \\
  &$\checkmark$& $\checkmark$ & 0.122 & 0.082 & 0.040 & 0.063 & 0.043 & 0.020  \\
 $\checkmark$ &$\checkmark$ & $\checkmark$ & \textbf{0.077} & \textbf{0.048} & \textbf{0.029} & \textbf{0.040} & \textbf{0.027} & \textbf{0.013} \\
\bottomrule
\end{tabular}
}
\caption{Ablation study of \textit{ReconDreamer-RL}, evaluating the effectiveness of each module on various metrics.}
\label{tab:recondreamer_rl}
\end{table*}

\begin{table*}[!h]
\centering
\setlength{\tabcolsep}{2pt} 
\resizebox{1\linewidth}{!}{
\fontsize{8}{9}\selectfont  
\begin{tabular}{cccccccccc}
\toprule
\multirow{2}{*}{Method} 
& \multicolumn{3}{c}{Lane Shift @ 3m} 
& \multicolumn{3}{c}{Lane Shift @ 6m} 
& \multicolumn{3}{c}{Lane Change} \\ 
\cmidrule(lr){2-4} \cmidrule(lr){5-7} \cmidrule(lr){8-10}
& NTA-IoU $\uparrow$ & NTL-IoU $\uparrow$ & FID $\downarrow$
& NTA-IoU $\uparrow$ & NTL-IoU $\uparrow$ & FID $\downarrow$
& NTA-IoU $\uparrow$ & NTL-IoU $\uparrow$ & FID $\downarrow$ \\
\midrule
w/o Video Diffusion Prior 
& 0.224 & 49.67 & 160.23 
& 0.148 & 44.89 & 256.42 
& 0.215 & 46.23 & 178.34 \\
w/ Video Diffusion Prior 
& \textbf{0.338} & \textbf{50.14} & \textbf{121.56} 
& \textbf{0.302} & \textbf{48.94} & \textbf{140.24} 
& \textbf{0.326} & \textbf{50.08} & \textbf{124.31} \\
\bottomrule
\end{tabular}
}
\caption{Impact of diffusion prior within ReconSimulator on novel view rendering quality across lane shift and lane change.}
\label{tab:nuscenes1}
\vspace{-1.5mm}
\end{table*}


\begin{table}[t]
\centering
\resizebox{1\linewidth}{!}{
\begin{tabular}{l c c c}
\toprule
Method & CR$\downarrow$ & DCR$\downarrow$ & SCR$\downarrow$ \\
\midrule
VAD~\cite{vad} & 0.449 & 0.293 & 0.156 \\
GenAD~\cite{genad} & 0.379 & 0.234 & 0.145 \\
VADv2~\cite{vadv2} & 0.436 & 0.276 & 0.160 \\
RAD~\cite{rad} & 0.317 & 0.210 & 0.107 \\
\textit{ReconDreamer-RL} & \textbf{0.089} & \textbf{0.053} & \textbf{0.036} \\
\bottomrule
\end{tabular}
}
\caption{Comparison of collision metrics in cut-in scenarios}
\label{tab:cornercases}
\end{table}



\subsection{Experimental Setup}

\paragraph{Implementation Details.} To demonstrate the capability of the \textit{ReconDreamer-RL} framework in enhancing training, we use it to train the end-to-end autonomous driving policy in RAD~\cite{rad}. The training process has two stages: the imitation learning stage and the reinforcement learning stage. In the first stage, the perception capabilities are trained, followed by planning pre-training through behavior cloning. In the second stage, reinforcement learning is employed within closed-loop 3DGS environments to optimize planning further. Finally, we build an evaluation benchmark including edited 3DGS environments and original 3DGS environments to test the performance of the end-to-end autonomous driving method. All 3DGS environments are reconstructed from  nuScenes~\cite{nuscenes}, and additional details on the edited scenes are in the supplement.

\paragraph{Baseline.} In comparative analysis, we select representative end-to-end autonomous driving methods. For imitation learning, we choose VAD~\cite{vad}, GenAD~\cite{genad}, and VADv2~\cite{vadv2}. For reinforcement learning, we use RAD~\cite{rad}, which is trained in closed-loop 3DGS environments.

\paragraph{Evaluation Metrics.}
Following RAD~\cite{rad}, we evaluate policy performance using six metrics. The Dynamic Collision Ratio (DCR) and Static Collision Ratio (SCR) quantify collisions with dynamic and static obstacles, and their sum is referred to as the Collision Ratio (CR). The Positional Deviation Ratio (PDR) and Heading Deviation Ratio (HDR) measure the ego vehicle’s deviation from the expert trajectory for position and heading, summed up and referred to as the Deviation Ratio (DR). To assess the realism of the reinforcement learning environment, we use three metrics from DriveDreamer4D~\cite{drivedreamer4d}: Novel Trajectory Agent IoU (NTA-IoU) for foreground quality, Novel Trajectory Lane IoU (NTL-IoU) for evaluating lane markings, and Fréchet Inception Distance (FID)~\cite{fid} to quantify overall rendering fidelity.

\begin{table}[h]
\centering
\resizebox{0.98\linewidth}{!}{
\begin{tabular}{cccc}
\toprule
 Simulator &  EmerNeRF & RAD-3DGS & ReconSimulator \\
\midrule
Speed (FPS) & 0.21 & 135 & 125 \\
\bottomrule
\end{tabular}
}
\caption{Comparison of rendering speeds among reinforcement learning simulators, including EmerNeRF~\cite{emernerf}, RAD~\cite{rad}, and ReconSimulator.}
\label{tab:simulator}
\end{table}

\subsection{Main Results}

\paragraph{Quantitative Results.} \textit{ReconDreamer-RL} enhances the performance of RAD~\cite{rad} across all evaluation metrics, as shown in Tab.~\ref{tab:comparison}. Traditional imitation learning-based methods, such as VAD~\cite{vad} and VADv2~\cite{vadv2}, struggle with closed-loop evaluation tasks due to the discrepancy between training and closed-loop inference. RAD~\cite{rad} enhances performance by incorporating reinforcement learning and training within a 3DGS-based simulator. However, due to limited corner case coverage and poor novel view rendering ability of 3DGS, RAD exhibits a higher collision rate in our challenging evaluation benchmark. However, \textit{ReconDreamer-RL} leverages diffusion prior and data augmentation to enhance end-to-end autonomous driving, achieving a 3$\times$ reduction in Collision Ratio compared to RAD~\cite{rad}.


\paragraph{Corner Case Results.} As shown in Tab.~\ref{tab:cornercases}, we compare the collision metrics of different methods in corner cases, focusing on cut-in situations. Imitation learning-based methods exhibit a high Dynamic Collision Ratio in cut-in corner cases, as such scenarios are completely absent from the training data. RAD~\cite{rad} improves the model's ability to handle corner cases through reinforcement learning, but still lacks the necessary data to train on such situations. In contrast, \textit{ReconDreamer-RL} improves the policy's performance through a two-stage process, achieving 404.5\% improvement in Collision Ratio over imitation methods.

\paragraph{Rendering Speed.} Rendering speed is crucial for reinforcement learning environments. As shown in Tab.~\ref{tab:simulator}, ReconSimulator achieves efficient rendering speed, thus satisfying the requirements of reinforcement learning tasks.


\paragraph{Qualitative Results.} As shown in Fig.~\ref{fig:visual}, we compare \textit{ReconDreamer-RL} and RAD~\cite{rad} under two corner cases. The first scenario involves a vehicle on the right of the ego car quickly cutting in. The second scenario features a vehicle on the right performing a cut-in, followed by a sudden brake. In the first case, \textit{ReconDreamer-RL} successfully avoids the collision and maintains a safe distance, whereas RAD~\cite{rad} fails to react. In the second case, although RAD~\cite{rad} attempts to avoid the collision by shifting lanes,  it fails to control its speed and steering angle properly. In contrast, \textit{ReconDreamer-RL} safely changes lanes and decelerates to avoid the collision.

\subsection{Ablation Studies}
\paragraph{ReconSimulator.}As shown in Tab.~\ref{tab:nuscenes1}, we conduct ablation experiments on ReconSimulator to demonstrate the impact of the diffusion prior on novel view rendering quality. Specifically, without ReconSimulator, the NTA-IoU and NTL-IoU drop by 54.46\% and 9.71\%, and the FID metrics increase by 104.43\% for the lane shift of 6m, indicating that the appearance of surrounding vehicles and lane markings has degraded substantially. Meanwhile, Tab.~\ref{tab:recondreamer_rl} shows that removing ReconSimulator leads to worse end-to-end autonomous driving training performance.

\paragraph{Impact of the DAA.}The DAA contributes to reducing collision occurrences by generating corner case scenarios to aid training. As demonstrated in Tab.~\ref{tab:recondreamer_rl}, introducing DAA decreases the Collision Ratio from 0.172 to 0.117.

\paragraph{Impact of the CTG.}In Tab.~\ref{tab:recondreamer_rl}, the benefit of CTG is validated by reductions in CR (from 0.172 to 0.143) and DR (from 0.073 to 0.052), indicating improved trajectory adherence. By addressing the data bias, CTG  enhances the action distribution learned during the imitation learning phase.

\section{Conclusion}
This paper introduces \textit{ReconDreamer-RL}, a framework designed to enhance reinforcement learning for end-to-end autonomous driving through the integration of video diffusion priors. It focuses on constructing a more realistic simulator, enriching corner cases, and addressing the issue of training data distribution. Specifically, we introduce ReconSimulator, which combines the video diffusion prior for appearance modeling and a kinematic model for physical dynamics, providing realistic environments for reinforcement learning. Additionally, we develop the DAA, which generates complex traffic scenarios, such as sudden lane changes. To address the issue of training data distribution being mostly on simple straight-line movements, CTG is proposed to enhance trajectory diversity through extension and interpolation. Experimental evaluations indicate that \textit{ReconDreamer-RL} significantly improves the performance of end-to-end autonomous driving models, reducing the Collision Ratio by 5$\times$ compared to imitation-based learning methods.




\bibliography{aaai2026}
\clearpage
\newpage

\twocolumn[
\begin{center}
\LARGE \textbf{Supplementary Material for ReconDreamer-RL: Enhancing Reinforcement Learning via Diffusion-based Scene Reconstruction}
\end{center}
\vspace{2cm}  
]

In the supplementary material, we provide detailed implementation information, including the datasets, metric calculations, training details, and the edited scenes along with their visualizations. Additionally, to demonstrate the generalization capability of \thename{}, we conduct further experiments on the Waymo~\cite{waymo} dataset and present additional visualizations comparing \thename{} with RAD~\cite{rad}. Finally, a video is provided that showcases the capabilities of \thename{}.

\section{Implementation Details}

\subsection{Datasets} 
We reconstruct the entire nuScenes~\cite{nuscenes} and Waymo~\cite{waymo} datasets into 3D Gaussian Splatting~(3DGS)~\cite{kerbl20233d} environments. These reconstructed environments will be made publicly available to support future research. 
\paragraph{nuScenes.} The nuScenes~\cite{nuscenes} dataset comprises a large collection of diverse urban driving scenarios. It provides about 1.4 million annotated 3D bounding boxes across 23 object categories. Scene imagery is captured using six cameras offering a full 360-degree horizontal field of view, and keyframes are annotated at 2 Hz.
\paragraph{Waymo.}To further evaluate the generalization capability of \thename{}, we also conduct experiments on the Waymo~\cite{waymo} dataset, a comprehensive benchmark featuring complex urban driving scenes under varied environmental conditions. It provides high-quality sensor data, including LiDAR, radar, and high-resolution imagery, enabling robust assessment across diverse scenarios.

\subsection{Metric}
To evaluate the performance of the autonomous driving policy, we use six key metrics following  RAD~\cite{rad}.  

The Dynamic Collision Ratio (DCR) represents the frequency of collisions with dynamic obstacles and is calculated as:
\begin{equation}
DCR = \frac{N_{\text{dc}}}{N_{\text{total}}},
\end{equation}
where \( N_{\text{dc}} \) denotes the number of clips where collisions with dynamic obstacles occur, and \( N_{\text{total}} \) represents the total number of clips.

The Static Collision Ratio (SCR) quantifies the occurrence of collisions with static obstacles and is defined as:
\begin{equation}
SCR = \frac{N_{\text{sc}}}{N_{\text{total}}},
\end{equation}
where \( N_{\text{sc}} \) is the number of clips with static obstacle collisions.

The Collision Ratio (CR) represents the overall collision frequency and is given by:
\begin{equation}
CR = DCR + SCR.
\end{equation}

The Positional Deviation Ratio (PDR) measures how closely the ego vehicle follows the expert trajectory in terms of position. It is defined as:
\begin{equation}
PDR = \frac{N_{\text{pd}}}{N_{\text{total}}},
\end{equation}
where \( N_{\text{pd}} \) denotes the number of trajectory clips in which the positional deviation exceeds a predefined threshold.

The Heading Deviation Ratio (HDR) evaluates orientation accuracy by calculating the proportion of clips in which the heading deviation exceeds a predefined threshold.
\begin{equation}
HDR = \frac{N_{\text{hd}}}{N_{\text{total}}},
\end{equation}
where \( N_{\text{hd}} \) is the number of clips where the heading deviation exceeds the threshold.  

The Deviation Ratio (DR) provides a combined metric of positional and orientation deviations from the expert trajectory, formulated as:  
\begin{equation}
DR = PDR + HDR.
\end{equation}

Meanwhile, since the edited scenes lack ground-truth expert trajectories, we only compute the collision-related metrics for these scenarios. And, we adopt Novel Trajectory Agent IoU (NTA-IoU) and Novel Trajectory Lane IoU (NTL-IoU) from DriveDreamer4D~\cite{drivedreamer4d} to evaluate the realism of the environment in reinforcement learning, as they are specifically designed to measure the spatiotemporal consistency of dynamic agents (foreground) and static lane structures (background) in rendered videos.

NTA-IoU applies the YOLO11 detector~\cite{khanam2024yolov11} to obtain 2D bounding boxes of vehicles from rendered images generated along novel trajectories. In parallel, the original 3D bounding boxes are transformed into the coordinate frame of the novel trajectory, allowing accurate projection into 2D bounding boxes under the new viewpoint. For each projected box, NTA-IoU identifies the closest detected box and computes the Intersection over Union (IoU).

In an analogous manner, the NTL-IoU is calculated by first applying geometric transformations to project lane structures from the original trajectories into the viewpoint of novel trajectories. Concurrently, lane lines in the newly rendered images are extracted using the TwinLiteNet model~\cite{che2023twinlitenet}. Ultimately, the mean Intersection over Union (IoU) between these projected lane lines and those detected by TwinLiteNet is computed.

\subsection{Training Details} 
To demonstrate the capability of the \thename{} framework in enhancing training, we apply it to train RAD~\cite{rad}. Next, we introduce the network architecture of RAD~\cite{rad} and its training stages. And, we summarize the detailed hyperparameter settings for both the imitation learning and reinforcement learning stages in Tab.~\ref{il} and  Tab.~\ref{rl}, respectively.

\paragraph{Network Architecture of RAD.} RAD~\cite{rad} takes multi-view images as input, transforms sensor data into scene token embeddings, and outputs the probabilistic distribution of actions. The framework consists of a BEV encoder, map head, agent head, image encoder, and planning head. Specifically, the BEV encoder transforms multi-view features from the perspective view into bird’s-eye-view (BEV) representations~\cite{li2024bevformer,yang2023bevformer}. The map head extracts vectorized road elements. The agent head predicts motion states and trajectories of surrounding dynamic agents. Finally, the planning head outputs the probabilistic distribution of driving actions. Meanwhile, RAD~\cite{rad} employs a two-stage training approach, including the imitation learning stage and the reinforcement learning stage.  
\paragraph{Imitation Learning Stage.}In this stage, perception training is conducted, where the map head and agent head explicitly predict map elements and agent motion information, supervised by ground-truth labels. Subsequently, imitation learning is employed to initialize the action distribution based on large-scale driving demonstrations from expert drivers~\cite{nuscenes,waymo,xiao2021pandaset}. 

\paragraph{Reinforcement Learning Stage.}In the reinforcement learning stage, independent 3DGS models are trained for various driving scenario video clips, serving as digital driving environments. The autonomous driving policy continuously interacts with these 3DGS environments, calculates rewards based on actions, and optimizes the driving policy using the PPO~\cite{ppo} algorithm.

\begin{table}[t]
\centering
\label{tab:planning_pre_training}
\resizebox{1\linewidth}{!}{
\begin{tabular}{l|c}
\textbf{Config}                         & \textbf{Imitation Learning}                   \\ \hline
Learning rate                           & $1 \times 10^{-4}$                               \\
Learning rate schedule                  & Cosine decay                                     \\
Optimizer                               & AdamW~\cite{loshchilov2017decoupled}            \\
Optimizer hyper-parameters              & $\beta_1=0.9$, $\beta_2=0.999$, $\epsilon=10^{-8}$ \\
Weight decay                            & $1 \times 10^{-4}$                               \\
Planning head dimension                 & 256                                             
\end{tabular}}
\caption{Hyperparameters for the Imitation Learning Stage.}
\label{il}
\end{table}

\begin{table}[t]
\centering
\label{tab:reinforced_post_training}
\resizebox{1\linewidth}{!}{
\begin{tabular}{l|c}
\textbf{Config}                         & \textbf{Reinforcement Learning}                   \\ \hline
Learning rate                           & $5 \times 10^{-6}$                                  \\
Learning rate schedule                  & Cosine decay                                        \\
Optimizer                               & AdamW~\cite{loshchilov2017decoupled}               \\
Optimizer hyper-parameters              & $\beta_1=0.9$, $\beta_2=0.999$, $\epsilon=10^{-8}$  \\
Weight decay                            & $1 \times 10^{-4}$                                  \\
GAE parameters                          & $\gamma=0.9$, $\lambda=0.95$                        \\
Clipping thresholds                     & $\epsilon_x=0.1$, $\epsilon_y=0.2$                  \\
Planning head dimension                 & 256                                                 
\end{tabular}}
\caption{Hyperparameters for the Reinforcement Learning Stage.}
\label{rl}
\end{table}

\begin{figure*}[t]
\centering
\setlength{\abovecaptionskip}{0.5em}
\includegraphics[width=1\linewidth]{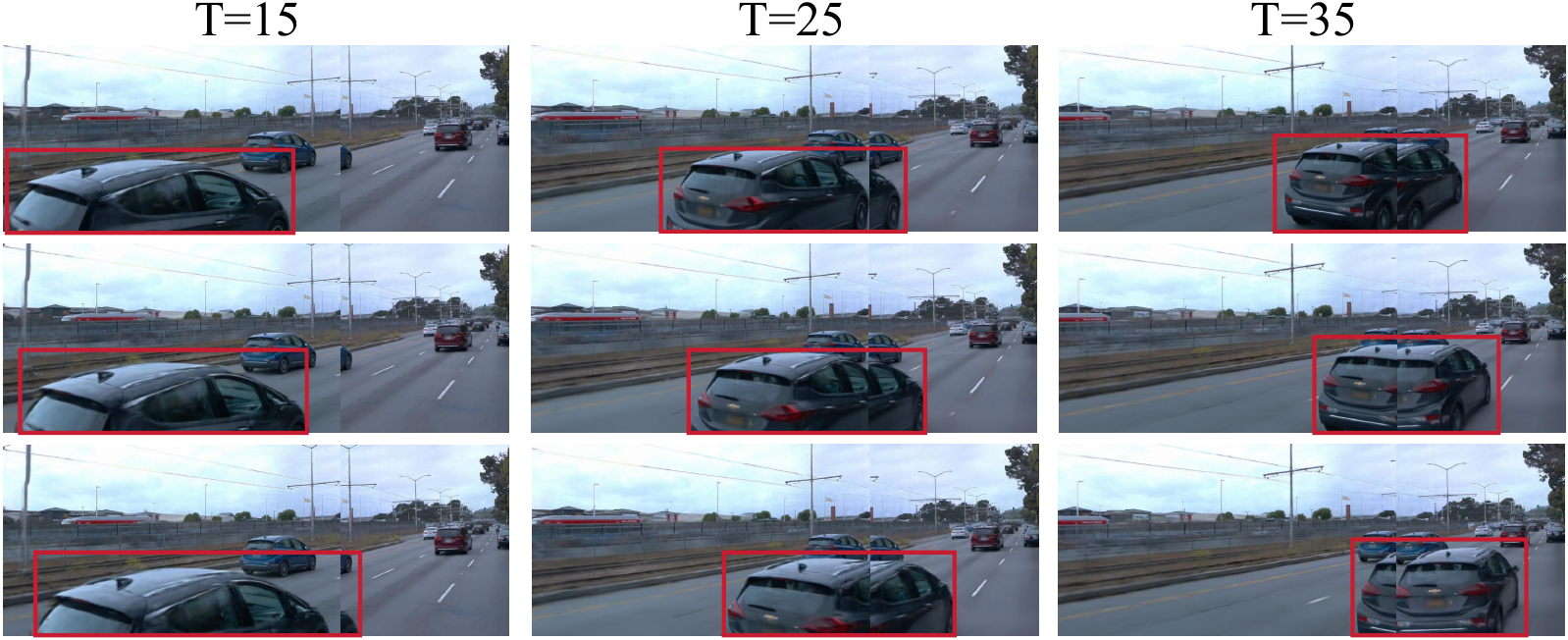}
\caption{Various trajectories generated by the Dynamic Adversary Agent (DAA) for the vehicle in the red box at the same timestep.}
\label{fig:DAA1} 
\end{figure*}

\begin{table*}[!h]
\centering
\setlength{\tabcolsep}{2pt} 
\resizebox{1\linewidth}{!}{
\fontsize{8}{9}\selectfont  
\begin{tabular}{cccccccccc}
\toprule
\multirow{2}{*}{Method} 
& \multicolumn{3}{c}{Lane Shift @ 3m} 
& \multicolumn{3}{c}{Lane Shift @ 6m} 
& \multicolumn{3}{c}{Lane Change} \\ 
\cmidrule(lr){2-4} \cmidrule(lr){5-7} \cmidrule(lr){8-10}
& NTA-IoU $\uparrow$ & NTL-IoU $\uparrow$ & FID $\downarrow$
& NTA-IoU $\uparrow$ & NTL-IoU $\uparrow$ & FID $\downarrow$
& NTA-IoU $\uparrow$ & NTL-IoU $\uparrow$ & FID $\downarrow$ \\
\midrule
w/o Video Diffusion Prior 
& 0.477 & 53.23  & 128.90
& 0.353 & 47.14 & 224.62 
& 0.486 & 55.34 &89.83 \\
w/ Video Diffusion Prior 
& \textbf{0.543} & \textbf{54.84} & \textbf{90.12}
& \textbf{0.454} & \textbf{52.92} & \textbf{148.23}
& \textbf{0.552} & \textbf{56.12} & \textbf{72.86} \\

\bottomrule
\end{tabular}
}
\caption{Impact of diffusion prior within ReconSimulator on novel view rendering quality across lane shift and lane change on the Waymo~\cite{waymo}.}
\label{tab:diffusionwaymo}
\end{table*}

\subsection{Details of Edited Scenes}
\begin{figure*}[t]
\centering
\setlength{\abovecaptionskip}{0.5em}
\includegraphics[width=0.8\linewidth]{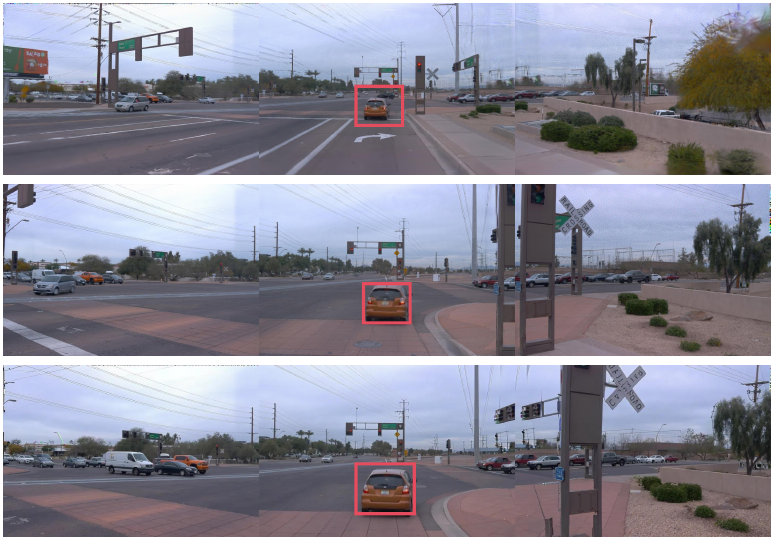}
\caption{Example scenario of BlockedIntersection. The vehicle in the red box is stationary and blocks the intersection when the ego vehicle intends to turn. The ego vehicle must brake or change lanes to the left to avoid a potential collision.}
\label{fig:cornercase2} 
\end{figure*}

\begin{figure*}[h]
\centering
\setlength{\abovecaptionskip}{0.5em}
\includegraphics[width=0.8\linewidth]{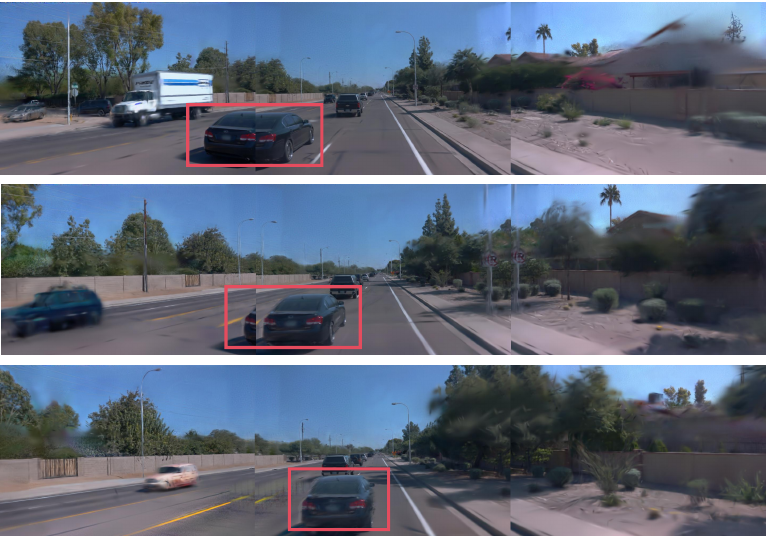}
\caption{Example scenario of DynamicCutIn. The vehicle in the red box is attempting to cut in from an adjacent lane while the ego vehicle is moving straight. The ego vehicle needs to slow down or brake to yield.}
\label{fig:cornercase5} 
\end{figure*}

\begin{figure*}[h]
\centering
\setlength{\abovecaptionskip}{0.5em}
\includegraphics[width=0.8\linewidth]{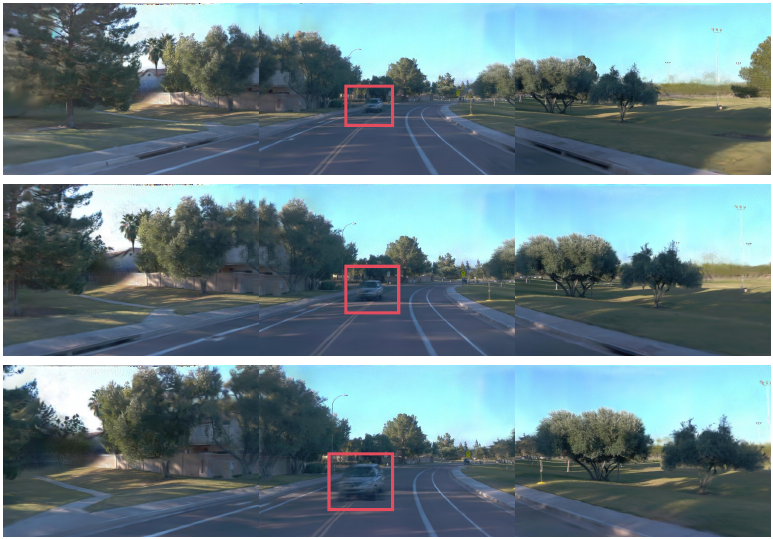}
\caption{Example scenario of OppositeLaneIntrusion. The vehicle in the red box is attempting to encroach into the ego vehicle's lane from the opposite direction. The ego vehicle must slow down or steer away to avoid a collision.}
\label{fig:cornercase4} 
\end{figure*}

\begin{figure*}[h]
\centering
\setlength{\abovecaptionskip}{0.5em}
\includegraphics[width=0.8\linewidth]{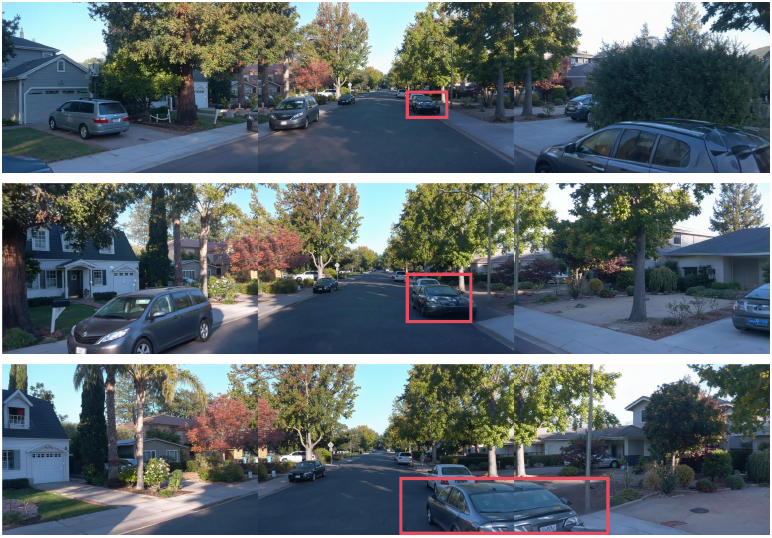}
\caption{Example scenario of ParkingCutIn. The vehicle in the red box is attempting to exit a parallel parking space and merge into the lane ahead. The ego vehicle must slow down or stop to avoid a collision.}
\label{fig:cornercase3} 
\end{figure*}

\begin{figure*}[h]
\centering
\setlength{\abovecaptionskip}{0.5em}
\includegraphics[width=0.8\linewidth]{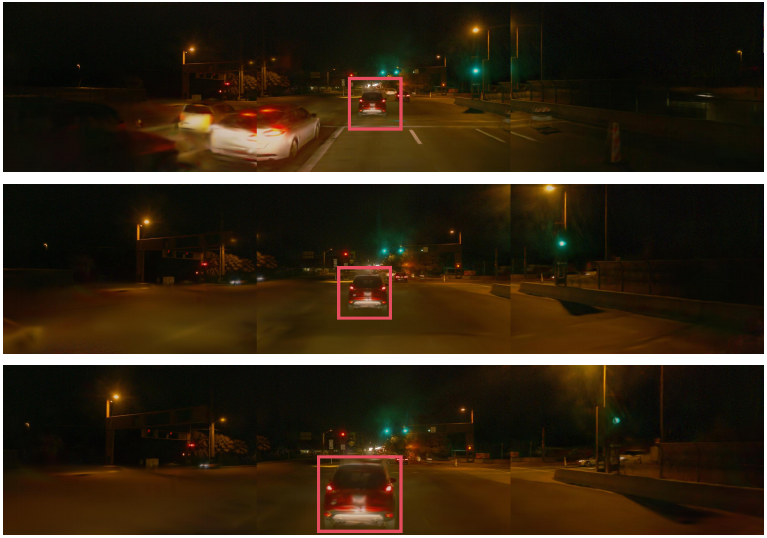}
\caption{Example scenario of HardBrake. The vehicle in the red box is decelerating abruptly. The ego vehicle needs to change lanes to avoid a rear-end collision.}
\label{fig:cornercase6} 
\end{figure*}

\begin{figure*}[h]
\centering
\setlength{\abovecaptionskip}{0.5em}
\includegraphics[width=0.8\linewidth]{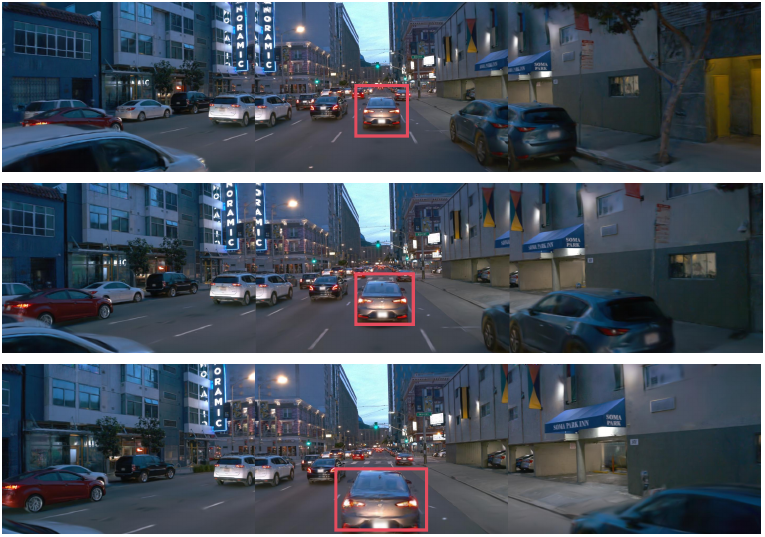}
\caption{Example scenario of HazardAtSideLane. The vehicle in the red box is slowly moving and partially occupying the ego vehicle's lane. The ego vehicle needs to change lanes to the left and merge into the traffic moving in the same direction.}
\label{fig:cornercase7} 
\end{figure*}

We design a series of corner cases inspired by~\cite{CARLA,jia2024bench2drive}. As shown in Fig.~\ref{fig:cornercase2} to~\ref{fig:cornercase7}, we present a diverse set of corner case scenarios, including BlockedIntersection, DynamicCutIn, OppositeLaneIntrusion, ParkingCutIn, HardBrake, and HazardAtSideLane. Then, we detail several corner cases from our edited scenes.
\paragraph{BlockedIntersection.} While performing a maneuver, the ego vehicle encounters a stationary vehicle blocking the road and must take evasive action or apply the brakes to avoid a collision. 
\paragraph{DynamicCutIn.} The ego vehicle must slow down or brake to allow a vehicle from traffic in an adjacent lane to cut in front. 
\paragraph{OppositeLaneIntrusion.} While driving straight, a vehicle from the opposite lane encroaches into the ego vehicle's lane, forcing the ego vehicle to brake or steer to the right to avoid a potential collision. 
\paragraph{ParkingCutIn.} The ego vehicle must reduce its speed or apply the brakes to allow a parked vehicle exiting a parallel parking space to merge into the lane ahead. 
\paragraph{HardBrake.} The leading vehicle decelerates abruptly, and the ego vehicle is required to perform an emergency stop. 
\paragraph{HazardAtSideLane.} The ego vehicle encounters a slow-moving obstacle partially blocking the lane and needs to maneuver into a lane of traffic moving in the same direction to bypass it. Unlike ParkedObstacle, this scenario places greater emphasis on merging into another lane. 
\paragraph{ParkedObstacle.} The ego vehicle encounters a parked car blocking part of the lane and must change lanes into moving traffic in the same direction to bypass it. 
\paragraph{MergeIntoSlowTraffic.} The ego vehicle is required to merge into a slow-moving traffic flow. 
\paragraph{Construction.} The ego vehicle encounters a construction zone blocking part of the lane and must shift into traffic moving in the same direction to navigate around it. In comparison to ParkedObstacle, the construction area occupies more of the lane's width, and the ego vehicle must temporarily divert from its planned route to bypass it. 
\paragraph{OppositeLaneRightTurn.} While preparing to make a right turn, a vehicle from the opposite lane enters the ego vehicle's lane, forcing it to move right to avoid a potential collision.
\paragraph{BlindIntersectionCrossing.} The ego vehicle approaches an intersection with its view obstructed by stationary objects or vehicles, requiring it to slow down or stop and proceed only after confirming safety.
\paragraph{WrongWayVehicle.} A vehicle traveling in the wrong direction approaches the ego vehicle's lane head-on, compelling it to immediately brake or steer away to avoid collision.
\paragraph{LaneChangeConflict.}
Both the ego vehicle and a vehicle in an adjacent lane simultaneously attempt to merge into the same lane, requiring the ego vehicle to yield or abort the lane change.
\paragraph{ParkedObstacleTwoWays.} The "TwoWays" version of ParkedObstacle. The ego vehicle faces a parked car blocking the lane and must change lanes into traffic moving in the opposite direction to avoid it. 
\paragraph{ConstructionTwoWays.} The "TwoWays" version of Construction. The ego vehicle encounters a construction area blocking the lane in both directions and must change lanes into traffic flowing in the opposite direction to avoid it. Compared to Construction, this obstruction occupies a wider lane area, requiring the ego vehicle to navigate through traffic in both directions. 
\paragraph{HazardAtSideLaneTwoWays.} The ego vehicle faces a slow-moving obstacle partially blocking the lane and must either brake or maneuver into traffic moving in the opposite direction to avoid it. 

\begin{table}[t]
\centering
\resizebox{1\linewidth}{!}{
\begin{tabular}{l c c c}
\toprule
Method & CR$\downarrow$ & DCR$\downarrow$ & SCR$\downarrow$ \\
\midrule
RAD~\cite{rad} & 0.140 & 0.093 & 0.046 \\
\thename{} & \textbf{0.093} & \textbf{0.060} & \textbf{0.033} \\
\bottomrule
\end{tabular}
}
\caption{Comparison of collision metrics on the Waymo~\cite{waymo} dataset}
\label{tab:waymo}
\end{table}

\begin{figure*}[t]
\centering
\setlength{\abovecaptionskip}{0.5em}
\includegraphics[width=1\linewidth]{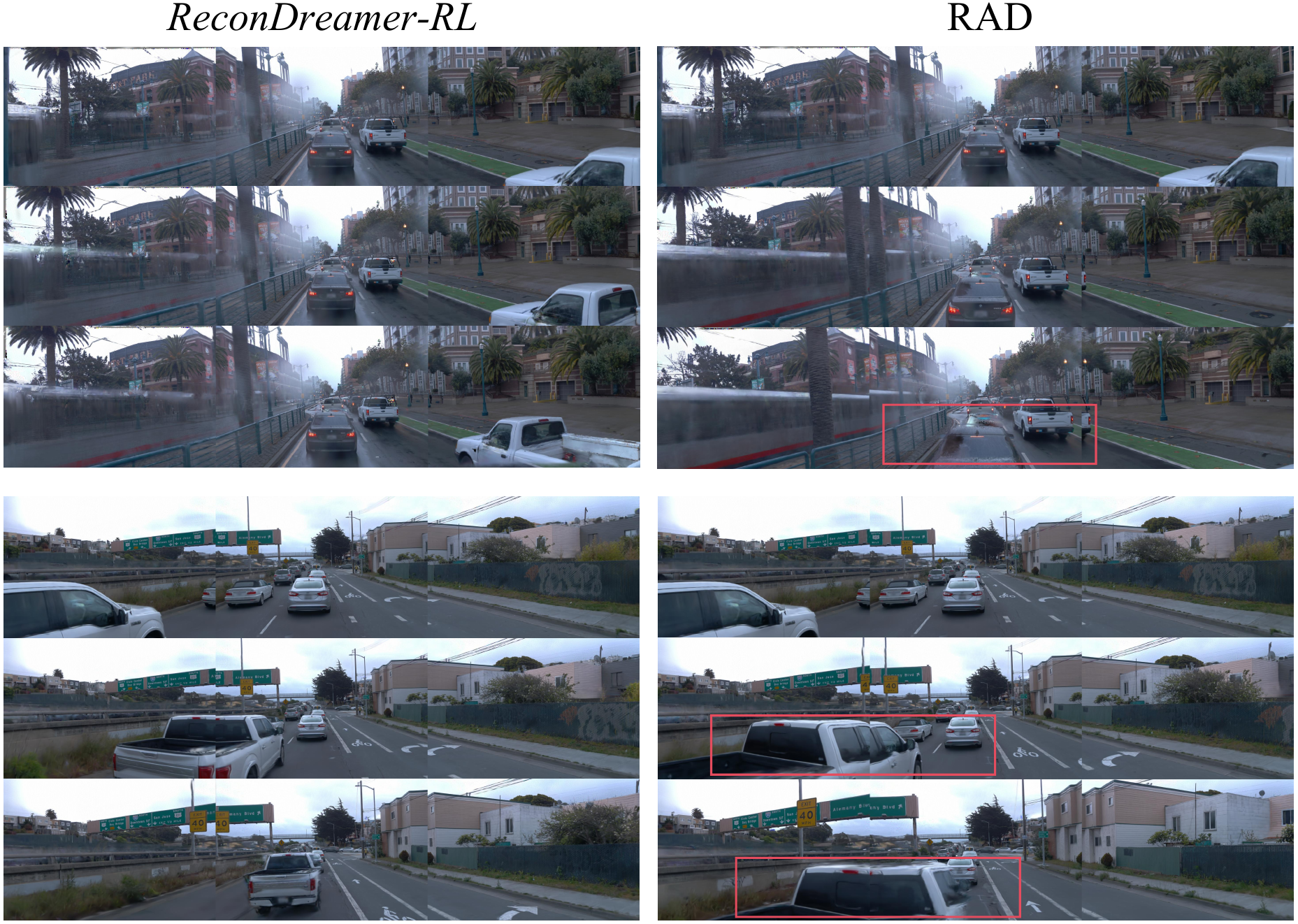}
\caption{Comparison of different methods in challenging corner cases, with collisions highlighted by red boxes.}
\label{fig:compare} 
\end{figure*}

\section{Additional Experimental Results.} 

\paragraph{ReconSimulator.} We also perform ablation experiments on ReconSimulator to assess the impact of the diffusion prior~\cite{recondreamer,drivedreamer2,drivedreamer} on novel view rendering quality, using the Waymo dataset~\cite{waymo}, as shown in Tab.~\ref{tab:diffusionwaymo}. Specifically, for the lane change, integrating the video diffusion prior significantly improves the NTA-IoU by 13.58\%, NTL-IoU by 1.41\%, and reduces the FID score by 18.89\%. These improvements clearly demonstrate that incorporating the video diffusion prior allows the rendered environment to provide better support for reinforcement learning tasks by enhancing the visual consistency and realism. Moreover, these results underline the strong generalization capability of ReconSimulator's appearance modeling across diverse datasets, highlighting its robustness and effectiveness across datasets.

\paragraph{Quantitative Results on Waymo.} As shown in Tab.~\ref{tab:waymo}, we quantitatively compare the performance of \thename{} and RAD~\cite{rad} on the Waymo dataset~\cite{waymo}. Specifically, \thename{} achieves a reduction in the Collision Ratio from 0.140 to 0.093, demonstrating superior safety performance and enhanced robustness in complex driving scenarios.

\paragraph{Qualitative Results on Waymo.}
As shown in Fig.~\ref{fig:compare}, we compare \thename{} and RAD~\cite{rad} under two corner cases: HardBrake and DynamicCutIn. \thename{} successfully avoids collisions by properly adjusting its speed, benefiting from a more realistic simulator enabled by diffusion-based scene reconstruction and further enhanced through data augmentation.

\paragraph{Visualization of DAA.} As mentioned in the main text, in the reinforcement learning stage, the Dynamic Adversary Agent (DAA) has a certain probability of fine-tuning the trajectories in the edited scenes (e.g., adjusting the target vehicle's speed) instead of directly reusing those generated during the imitation stage. In Fig.~\ref{fig:DAA1}, we present different trajectories generated by DAA at the same timestep. Since the method directly modifies the speed based on the original trajectory, it does not affect the overall timing of the pipeline.


\newpage

\end{document}